%% file: arxiv.tex
\documentclass{article}

\usepackage{microtype}
\usepackage{graphicx}
\usepackage{subfigure}
\usepackage{booktabs} 

\usepackage{hyperref}

\usepackage[accepted]{icml2025}

\usepackage{amsmath}
\usepackage{amssymb}
\usepackage{mathtools}
\usepackage{amsthm}

\usepackage[capitalize,noabbrev]{cleveref}

\usepackage{hyperref}
\usepackage{url}
\usepackage{hyperref}
\usepackage{graphics}
\usepackage[utf8]{inputenc} 
\usepackage[T1]{fontenc}    
\usepackage{hyperref}       
\usepackage{url}            
\usepackage{booktabs}       
\usepackage{amsfonts}       
\usepackage{nicefrac}       
\usepackage{microtype}      
\usepackage{xcolor}         
\usepackage[utf8]{inputenc}
\usepackage{mathtools}
\usepackage{amsthm}
\usepackage{arydshln}
\usepackage{multirow}
\usepackage{wrapfig, lipsum, booktabs}
\usepackage{paralist, tabularx}
\usepackage{balance}
\usepackage{pgfplots}
\usetikzlibrary{pgfplots.groupplots}
\pgfplotsset{compat=1.3}
\usepackage{tikz}
\usetikzlibrary{patterns}
\usepackage{pgf-pie}
\usepackage{adjustbox}
\usepackage{colortbl}
\usepackage{subcaption}
\usepackage{xspace}
\usepackage{enumitem}

\theoremstyle{plain}

\theoremstyle{definition}

\theoremstyle{remark}

\usepackage[textsize=tiny]{todonotes}

\makeatletter

\newcommand{\Rmnum}[1]{\expandafter\@slowromancap\romannumeral #1@}
\makeatother

\definecolor{my_yellow}{RGB}{248,236,198}
\definecolor{my_green}{RGB}{197,224,180}

\definecolor{demphcolor}{RGB}{144, 144, 144}
\newcommand{\demph}[1]{\textcolor{demphcolor}{#1}}

\newcommand{\acronym}[1]{\underline{\textbf{#1}}}
\newcommand{\modelname}{NBP\xspace}

\icmltitlerunning{Next Block Prediction: Video Generation via Semi-Autoregressive Modeling}

\begin{document}

\twocolumn[
\icmltitle{Next Block Prediction: Video Generation via Semi-Autoregressive Modeling}



\icmlsetsymbol{equal}{*}

\begin{icmlauthorlist}
\icmlauthor{Shuhuai Ren}{1}
\icmlauthor{Shuming Ma}{2}
\icmlauthor{Xu Sun}{1}
\icmlauthor{Furu Wei}{2}
\end{icmlauthorlist}

\icmlaffiliation{1}{National Key Laboratory for Multimedia Information Processing, School of Computer Science, Peking University}
\icmlaffiliation{2}{Microsoft Research}

\icmlcorrespondingauthor{Xu Sun}{xusun@pku.edu.cn}
\icmlcorrespondingauthor{Furu Wei}{fuwei@microsoft.com}

\hypersetup{urlcolor=black}
\begin{center}
\url{https://renshuhuai-andy.github.io/NBP-project/}
\end{center}

\icmlkeywords{Machine Learning, ICML}

\vskip 0.3in
]



\printAffiliationsAndNotice{}  

\begin{abstract}
Next-Token Prediction (NTP) is a de facto approach for autoregressive (AR) video generation, but it suffers from suboptimal unidirectional dependencies and slow inference speed.  
In this work, we propose a semi-autoregressive (semi-AR) framework, called \acronym{N}ext-\acronym{B}lock \acronym{P}rediction (\modelname), for video generation. 
By uniformly decomposing video content into equal-sized blocks (e.g., rows or frames), we shift the generation unit from individual tokens to blocks, allowing each token in the current block to simultaneously predict the corresponding token in the next block. 
Unlike traditional AR modeling, our framework employs bidirectional attention within each block, enabling tokens to capture more robust spatial dependencies. By predicting multiple tokens in parallel, \modelname models significantly reduce the number of generation steps, leading to faster and more efficient inference. 
Our model achieves FVD scores of 103.3 on UCF101 and 25.5 on K600, outperforming the vanilla NTP model by an average of 4.4. 
Furthermore, thanks to the reduced number of inference steps, the \modelname model generates 8.89 frames (128$\times$128 resolution) per second, achieving an 11$\times$ speedup. We also explored model scales ranging from 700M to 3B parameters, observing significant improvements in generation quality, with FVD scores dropping from 103.3 to 55.3 on UCF101 and from 25.5 to 19.5 on K600, demonstrating the scalability of our approach.
\end{abstract}

\section{Introduction}
The advance of Large Language Models (LLMs) such as ChatGPT~\citep{chatgpt}, GPT-4~\citep{Achiam2023GPT4TR} and LLaMA~\citep{Touvron2023LLaMAOA} has cemented the preeminence of Autoregressive (AR) modeling in the realm of natural language processing (NLP). This AR modeling approach, combined with the decoder-only Transformer architecture~\citep{Vaswani2017AttentionIA}, has been pivotal in achieving advanced levels of linguistic understanding, generation, and reasoning~\citep{wei2022emergent, gpt-o1, chen-etal-2024-pca}. 
Recently, there is a growing interest in extending AR modeling from language to other modalities, such as images and videos, to develop a unified multimodal framework~\citep{gpt4o, Team2024ChameleonME, Lu2023UnifiedIO2S, Wu2023NExTGPTAM, chen2024next}. Such an AR-based framework brings numerous benefits: (1) It allows for the utilization of the well-established infrastructure and learning recipes from the LLM community~\citep{Dao2022FlashAttentionFA, kwon2023efficient}; (2) The scalability and generalizability of AR modeling, empirically validated in LLMs~\citep{Kaplan2020ScalingLF, Yu2023ScalingAM}, can be extended to the multimodal domains to strengthen models~\citep{henighan2020scaling}; (3) Cognitive abilities observed in LLMs can be transferred and potentially amplified with multimodal data, moving closer to the goal of artificial general intelligence~\citep{bubeck2023sparks}.

Given the inherently autoregressive nature of video data in temporal dimensions, video generation is a natural area for extending AR modeling. 
Vanilla AR methods for video generation typically follow the Next-Token Prediction (NTP) approach, i.e., tokenize video into discrete tokens, then predict each subsequent token based on the previous ones. 
However, this approach has notable limitations. First, the generation order of NTP often follows a unidirectional raster-scan pattern~\citep{hong2022cogvideo, Wang2024OmniTokenizerAJ, Yan2021VideoGPTVG}, which fails to capture strong 2D correlations within video frames, limiting the modeling of spatial dependencies~\citep{Tian2024VisualAM}. Second, NTP necessitates a significant number of forward passes during inference (e.g., 1024 steps to generate a 16-frame clip), which reduces efficiency and increases the risk of error propagation~\citep{bengio2015scheduled}.

In this work, we propose a semi-autoregressive (semi-AR) framework, called \acronym{N}ext-\acronym{B}lock \acronym{P}rediction (\modelname), for video generation. 
To better model local spatial dependencies and improve inference efficiency, our framework shifts the generation unit from individual tokens to blocks (e.g., rows or frames). The objective is also redefined from next-token to next-block prediction, where each token in the current block simultaneously predicts the corresponding token in the next block. 
In contrast to the vanilla AR framework, which attends solely to prefix tokens, our \modelname approach allows tokens to attend to all tokens within the same block via bidirectional attention, thus capturing more robust spatial relationships. By predicting multiple tokens in parallel, \modelname models significantly reduce the number of generation steps, resulting in faster and more computationally efficient inference. 

Experimental results on the UCF-101~\citep{Soomro2012UCF101AD} and Kinetics-600 (K600)~\citep{Carreira2018ASN} datasets demonstrate the superiority of our semi-AR framework. With the same model size (700M parameters), \modelname achieves a 103.3 FVD on UCF101 and a 25.5 FVD on K600, surpassing the vanilla NTP model by 4.4. Additionally, due to the reduced number of inference steps, \modelname models can generate 8.89 frames (128$\times$128 resolution) per second, achieving an 11$\times$ speedup in inference. Compared to previous state-of-the-art token-based models, our approach proves to be the most effective. Scaling experiments with models ranging from 700M to 3B parameters show a significant improvement in generation quality, with FVD scores dropping from 103.3 to 55.3 on UCF101 and from 25.5 to 19.5 on K600, highlighting the scalability of the framework. We hope this work inspires further advancements in the field.

\begin{figure*}[htbp]
    \centering
    \begin{minipage}[b]{0.39\textwidth}
        \centering
        \includegraphics[width=.9\textwidth]{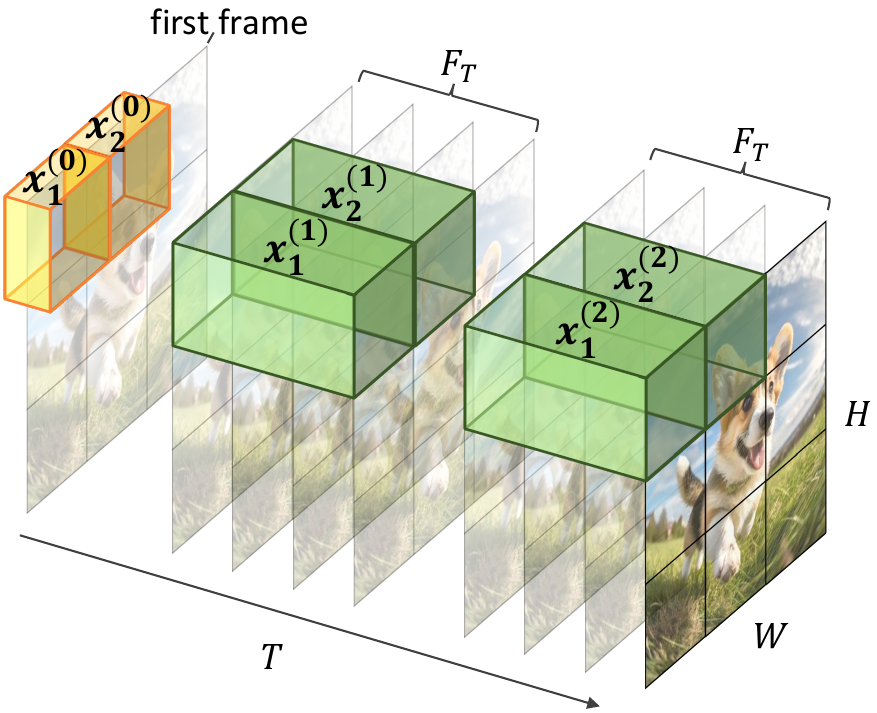}
        \caption{3D discrete token map produced by our video tokenizer. The input video consists of \colorbox{my_yellow}{one initial frame}, followed by $n$ \colorbox{my_green}{clips}, with each clip containing $F_T$ frames. $x^{(i)}_{j}$ indicates the $j^{th}$ video token in the $i^{th}$ clip.}
        \label{fig:tokenizer}
    \end{minipage}
    \hfill
    \begin{minipage}[b]{0.59\textwidth}
        \centering
        \includegraphics[width=\textwidth]{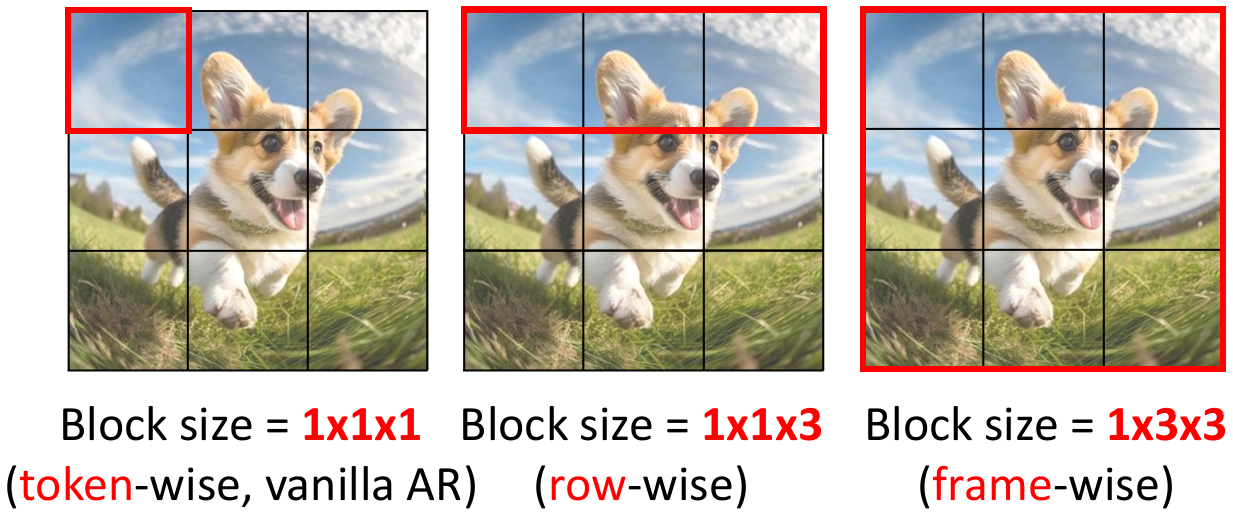}
        \caption{Examples of block include token-wise, row-wise, and frame-wise representations. When the block size is set to 1$\times$1$\times$1, it degenerates into a token, as used in vanilla AR modeling. Note that the actual token corresponds to a 3D cube, we omit the time dimension here for clarity.}
        \label{fig:block_example}
    \end{minipage}
\end{figure*}

\section{Related Work}
\paragraph{Video Generation.}
Prevalent video generation frameworks in recent years include Generative Adversarial Networks (GANs)~\citep{Yu2022GeneratingVW, Skorokhodov2021StyleGANVAC}, diffusion models~\citep{Ho2022ImagenVH, Ge2023PreserveYO, Gupta2023PhotorealisticVG, Yang2024CogVideoXTD}, autoregressive models~\citep{hong2022cogvideo, Yan2021VideoGPTVG, Kondratyuk2023VideoPoetAL}, etc. 
GANs can generate videos with rich details and high visual realism, but their training is often unstable and prone to mode collapse. In contrast, diffusion models exhibit more stable training processes and typically produce results with greater consistency and diversity~\citep{Yang2022DiffusionMA}. 
Nevertheless, AR models demonstrate significant potential for processing multi-modal data (e.g., text, images, audio, and video) within a unified framework, offering strong scalability and generalizability. To align with the trend of natively multimodal development~\citep{gpt4o}, this paper focuses on exploring video generation using AR modeling.

\paragraph{Autoregressive Models for Video Generation.}
With the success of the GPT series models~\citep{Brown2020LanguageMA}, a range of studies has applied AR modeling to both image~\citep{Chen2020GenerativePF, Lee2022AutoregressiveIG, wang2024loong, pang2024randar} and video generation~\citep{hong2022cogvideo, Wang2024OmniTokenizerAJ, Yan2021VideoGPTVG}. 
For image generation, traditional methods divide an image into a sequence of tokens following a raster-scan order and then predict each subsequent token based on the preceding ones. In video generation, this process is extended frame by frame to produce temporally-coherence content. 
However, conventional AR models predict only one token at a time, resulting in a large number of forward steps during inference. This significantly impairs the generation speed, especially for high-resolution images or videos containing numerous tokens~\citep{liu2024lumina-mgpt}.

\paragraph{Semi-Autoregressive Models.}
To improve the efficiency of AR models, early NLP researchers has explored semi-autoregressive modeling by generating spans of tokens instead of individual tokens per step~\citep{wang2018semi}. However, due to the variable length of text generation targets, it is challenging to predefine span sizes. Furthermore, fixed-length spans can disrupt semantic coherence and completeness, leading to significant degradation in generation quality; for instance, using a span length of 6 results in a 12\% drop in performance for English-German translation tasks~\citep{wang2018semi}.
More advanced semi-AR approaches, such as parallel decoding~\citep{Stern2018BlockwisePD} and speculative decoding~\citep{Xia2022SpeculativeDE}, typically use multiple output heads or additional modules (e.g., draft models) to predict several future tokens based on the last generated token~\citep{Gu2017NonAutoregressiveNM, Gloeckle2024BetterF}. 
In the context of video, where content can be uniformly decomposed into equal-sized blocks (e.g., row by row or frame by frame), we propose a framework where each token in the last block predicts the corresponding token in the next block, without requiring additional heads or modules. 


\paragraph{Multi-token Prediction in Image Generation.}
Recent work in the image generation field has also shown a pattern of multi-token prediction, albeit with different motivations and approaches. 
For example, VAR~\citep{Tian2024VisualAM} employs a coarse-to-fine strategy across resolution scales, whereas our method processes spatiotemporal blocks at original resolution, achieving over 2$\times$ token efficiency (256 vs. 680 tokens for a 256$\times$256 frame). 
Unlike MAR~\citep{Li2024AutoregressiveIG}, which relies on randomized masking (70\% mask rate) and suffers from partial supervision (30\% of unmasked tokens do not receive supervision), our approach eliminates mask token modeling entirely, ensuring full supervision and improved training efficiency. 
While other works explore specialized token combinations~\citep{li2023lformer,wang2024parallelized}, our method minimizes architectural priors, enabling seamless adaptation from pre-trained NTP models and superior performance, especially for video generation.

\section{Method}
In this section, we first introduce our video tokenizer $\S$~\ref{subsec:video-tokenization}, highlighting its two key features: joint image-video tokenization and temporal causality, both of which facilitate our semi-AR modeling approach. 
Next, we provide a detailed comparison between vanilla Next-Token Prediction (NTP) ($\S$~\ref{subsec:ar}) and our \acronym{N}ext-\acronym{B}lock \acronym{P}rediction (\modelname) modeling ($\S$~\ref{subsec:semi-ar}). 
Our \modelname framework employs a block-wise objective function and attention masking, enabling more efficient capture of spatial dependencies and significantly improving inference speed.

\subsection{Preliminary I: Video Tokenization}
\label{subsec:video-tokenization}
We reproduce closed-source MAGVITv2~\cite{yu2023language} as our video tokenizer, which is based on a causal 3D CNN architecture. 
Given a video $\mathbf{X} \in \mathbb{R}^{T \times H \times W \times 3}$ in RGB space,\footnote{Images can be considered as ``static'' videos with $T=1$.} MAGVITv2 encodes it into a feature map $\mathbf{Z} \in \mathbb{R}^{T' \times H' \times W' \times d}$, where $(T', H', W')$ is the latent size of $\mathbf{Z}$, and $d$ is the hidden dimension of its feature vectors.
After that, we apply a quantizer to convert this feature map $\mathbf{Z}$ into a discrete tokens map $\mathbf{Q} \in \mathbb{V}^{T' \times H' \times W'}$ (illustrated in Fig.~\ref{fig:tokenizer}), where $\mathbb{V}$ represents a visual vocabulary of size $|\mathbb{V}|=K$. 
After tokenization, these discrete tokens $\mathbf{Q}$ can be passed through a causal 3D CNN decoder to reconstruct the video $\hat{\mathbf{X}}$. 
We note that MAGVITv2 has two major advantages:

\paragraph{(1) Joint Image-Video Tokenization.} MAGVITv2 allows tokenizing images and videos with a shared vocabulary. 
To achieve this, the number of frames in an input video, $T$, must satisfy $T=1+n \times F_T$, meaning the video comprises an initial frame followed by $n$ clips, each containing $F_T$ frames. 
When $n=0$, the video contains only the initial frame, thus simplifying the video to an image. 
Both the initial frame and each subsequent clip are discretized into a $(1, H', W')$ token map. Therefore, the latent temporal dimension $T'$ of the token map $\mathbf{Q}$ equals to $1+n$, which achieves $F_T$ times downsampling ratio on the temporal dimension (except for the first frame). 
Additionally, $H' = \frac{H}{F_H}$ and $W' = \frac{W}{F_W}$, where $F_H, F_W$ are spatial downsampling factors.

\paragraph{(2) Temporal Causality.} The causal 3D CNN architecture ensures that the tokenization and detokenization of each clip depend only on the preceding clips, facilitating autoregressive modeling along the temporal dimension, which will be discussed further in $\S$~\ref{subsec:semi-ar}.

\begin{figure*}[tbp]
\centering
\includegraphics[width=.9\textwidth]{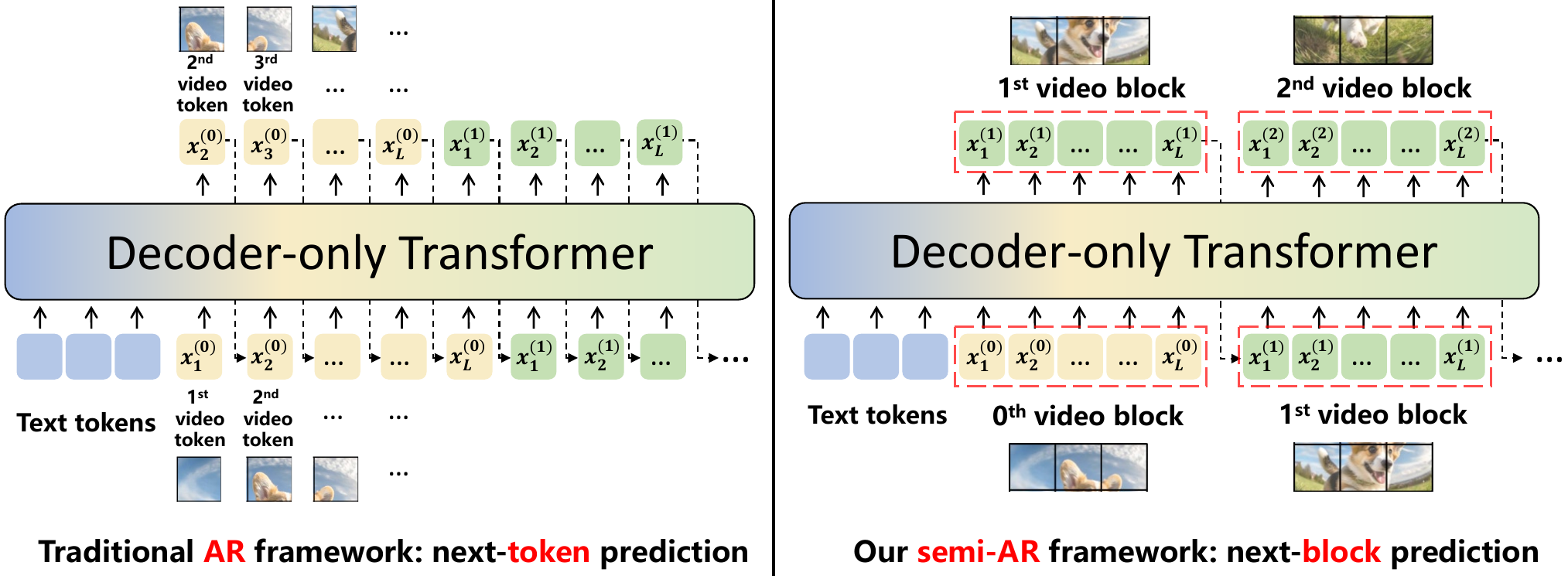}
\caption{Comparison between a vanilla autoregressive (AR) framework based on next-token prediction (left) and our semi-AR framework based on next-block prediction (right). $x^{(i)}_{j}$ indicates the $j^{th}$ video token in the $i^{th}$ block, with each block containing $L$ tokens. 
The dashed line in the right panel presents that the $L$ tokens generated in the current step are duplicated and concatenated with prefix tokens, forming the input for the next step's prediction during inference.}
\label{fig:framework}
\end{figure*}

\subsection{Preliminary II: Autoregressive Modeling for Video Generation}
\label{subsec:ar}
Inspired by the success of AR models in the field of NLP, previous work~\citep{Yan2021VideoGPTVG, Wu2021GODIVAGO, Wu2021NWAVS} has extended AR models to video generation. Typically, these methods flatten the 3D video token input $\mathbf{Q} \in \mathbb{V}^{T' \times H' \times W'}$ into a 1D token sequence. 
Let \colorbox{my_green}{$C^{(t)}=\{x^{(t)}_1, x^{(t)}_{2}, \dots, x^{(t)}_{L}\}$} be the set of tokens in the $t^{th}$ clip, where $L = H' \times W' = |C^{(t)}|$ is the total number of tokens in each clip, and every clip contains an equal number of tokens. 
Specially, when $t=0$, \colorbox{my_yellow}{$C^{(0)}$} denotes the first frame's tokens. 
Therefore, the 1D token sequence can be represented as 
$($
\colorbox{my_yellow}{$C^{(0)}$}
$, \dots,$
\colorbox{my_green}{$C^{(T')}$}
$)=($
\colorbox{my_yellow}{$x^{(0)}_1, x^{(0)}_2, \dots,x^{(0)}_L$}
$, \dots,$
\colorbox{my_green}{$x^{(T')}_1, x^{(T')}_2, \dots, x^{(T')}_L$}
$)$. 
In the AR framework, the next-token probability is conditioned on the preceding tokens, where each token $x^{(t)}_l$ depends only on its prefix $(x^{(<t)}_l, x^{(t)}_{<l})$. This unidirectional dependency allows the likelihood of the 1D sequence to be factorized as: 
\begin{equation}
\label{eq:ar}
p\left(x^{(0)}_1, \dots, x^{(T')}_L \right)
=\prod_{t=1}^{T'} \prod_{l=1}^{L} p\left(x^{(t)}_l \mid x^{(<t)}_l, x^{(t)}_{<l} \right)
\end{equation}
Since only one token is predicted per step, the inference process can become computationally expensive and time-consuming~\citep{liu2024lumina-mgpt}, motivating the exploration of more efficient methods, such as semi-AR models~\citep{wang2018semi}, to improve both speed and scalability.

\subsection{Semi-AR Modeling via Next Block Modeling}
\label{subsec:semi-ar}
In contrast to text, which consists of variable-length words and phrases, video content can be uniformly decomposed into equal-sized blocks (e.g., rows or frames). Fig.~\ref{fig:block_example} shows examples of token-wise, row-wise, and frame-wise block representations. 
Based on this, we propose a semi-autoregressive (semi-AR) framework named \acronym{N}ext-\acronym{B}lock \acronym{P}rediction (\modelname), where each token in the current block predicts the corresponding token in the next block. 
Fig.~\ref{fig:framework} illustrates an example of next-clip prediction, where each clip is treated as a block, and the next clip is predicted based on the preceding clips. 
This approach introduces two key differences compared to vanilla NTP modeling: 
\textbf{(1) Change in the generation target.} In \modelname, the $l^{th}$ token $x_l^{(t)}$ in the $t^{th}$ clip predicts $x_l^{(t+1)}$ in the next clip, rather than $x_{l+1}^{(t)}$ as in NTP. 
\textbf{(2) Increase in the number of generation targets.} Instead of predicting one token at a time, all $L$ tokens $x_{1:L}^{(t)}$ simultaneously predict the corresponding $L$ tokens $x_{1:L}^{(t+1)}$ in the next clip.
Accordingly, the \modelname objective function can be expressed as: 
\begin{equation}
\label{eq:semi-ar}
p\left(x^{(0)}_1, \ldots, x^{(T')}_L \right) 
= \prod_{t=1}^{T'} p\left( \colorbox{my_green}{$x_{1:L}^{(t)}$} \mid \colorbox{my_yellow}{$x_{1:L}^{(0)}$}, \ldots, \colorbox{my_green}{$x_{1:L}^{(t-1)}$} \right)
\end{equation}
By adjusting the block size, the framework can generate videos using different generation units. To ensure the effectiveness of this approach, four key components are designed:

\paragraph{(1) Initial Condition.}
In NTP models, a special token (e.g., \texttt{[begin\_of\_video]}) is typically used as the initial condition. In the \modelname setting, we can introduce a block of special tokens to serve as the initial condition for generating the first block. 
However, our preliminary experiments revealed that learning the parallel generation from the special token block to the first block is quite challenging. To address this issue, we propose two methods:
\textbf{(i) Taking the first frame $C^{(0)}$ as the initial condition.} In practice, following~\citet{girdhar2023emu}, users can upload an image as the first frame, or call an off-the-shelf text-to-image model (e.g., SDXL~\citep{podell2023sdxl}) to generate it. 
\textbf{(ii) Adopting a hybrid generation process}~\citep{wang2024parallelized}. Specifically, we can use per-token AR generation for the tokens in the first block. After the first block is generated, we then shift to per-block semi-AR generation. 
In order to make a fair comparison with other baselines, we used method (ii) in our experiments rather than relying on an extra first frame. 
Lastly, we note that both NTP and \modelname models can accept various inputs (e.g., text) as additional conditions (see Fig.~\ref{fig:framework}).

\begin{figure}
    \centering
    \includegraphics[width=.9\linewidth]{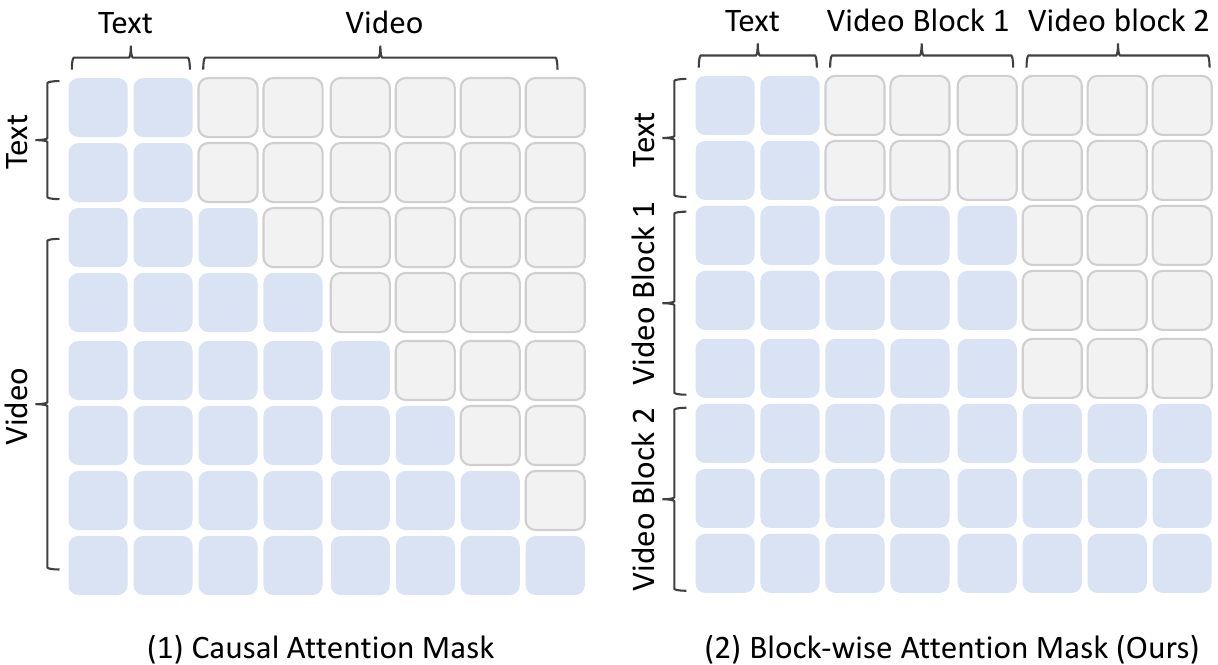}
    \caption{Causal attention mask in NTP modeling v.s. block-wise attention mask in \modelname modeling. The x-axis and y-axis represent keys and queries, respectively.}
\label{fig:attn-mask}
\end{figure}

\paragraph{(2) Block-wise Attention.}
To better capture spatial dependency, we allow tokens to attend to all tokens within the same block via bidirectional attention. Fig.~\ref{fig:attn-mask} compares traditional causal attention in NTP modeling with block-wise attention in \modelname modeling. 

\paragraph{(3) Block Size and Block Shape.}
The size and shape of blocks significantly influence generation quality, prompting us to conduct a comprehensive ablation study in 
$\S$~\ref{subsec:ablation} to identify the optimal configuration. 
Generally, we exclude blocks that span multiple frames (block shape with $T>1$) for several reasons:
\textbf{(i) Temporal Compression Constraints}: Input videos are sampled at 8 FPS or 16 FPS and undergo 4$\times$ temporal downsampling during tokenization, resulting in substantial information compression along the temporal dimension. Modeling rapidly changing content simultaneously across frames presents considerable challenges. 
\textbf{(ii) Causal Temporal Dynamics}: Our goal for the \modelname framework is not only to excel in video generation but also to serve as a potential world model~\citep{bruce2024genie, ha2018world}. Since videos represent the world in spatiotemporal dimensions and temporal changes are inherently causal, we aim to preserve complete causality in the temporal dimension during generation. Using a block shape with $T=1$ avoids introducing bidirectional temporal attention, aligning with our philosophy of employing an autoregressive generator (a decoder-only transformer) and a tokenizer like MagVITv2 with $T=1$ as the temporal unit. Results in Table~\ref{tab:block_shape} confirm that the block shape with $T=1$ achieve superior model performance.

\paragraph{(4) Inference Process.}
To illustrate the inference process of next-block prediction, we consider a scenario where each block corresponds to a clip. As shown in the right panel of Fig.~\ref{fig:framework}, during inference, the last $L$ tokens of the current output represent the predicted tokens for the next block.
These tokens are retained and concatenated with clip prefix, forming the input for the next step.  
By transitioning from token-by-token to block-by-block prediction, the \modelname framework leverages parallelization, reducing the number of generation steps by a factor of $L$, thereby decreasing computational cost and accelerating inference.


\section{Experiments}
\subsection{Experimental Setups}

\paragraph{Video Tokenizer.}
As MAGVITv2 is not open-sourced, we implemented it based on the original paper. In contrast to the official implementation, which utilizes LFQ~\citep{yu2023language} as its quantizer, we adopt FSQ~\citep{Mentzer2023FiniteSQ} due to its simplicity and reduced number of loss functions and hyper-parameters. Following the original paper's recommendations, we set the FSQ levels to $[8, 8, 8, 5, 5, 5]$, and the size of the visual vocabulary is 64K. 
Moreover, we employ PatchGAN~\citep{Isola2016ImagetoImageTW} instead of StyleGAN~\citep{Karras2018ASG} to enhance training stability.  
The reconstruction performance of our tokenizer is presented in Table~\ref{tab:video_reconstruction}, and additional training details are available in Appendix~\ref{app:model}. We note that MAGVITv2 is not open-sourced, we have made every effort to replicate its results. 
Our tokenizer surpasses OmniTokenizer~\cite{Wang2024OmniTokenizerAJ}, MAGVITv1~\cite{yu2023magvit}, and other models in performance. However, due to limited computational resources, we did not pre-train on ImageNet~\citep{Russakovsky2014ImageNetLS} or employ a larger visual vocabulary (e.g., 262K as in the original MAGVITv2), which slightly impacts our results compared to the official MAGVITv2. 
Nevertheless, we note that the primary objective of this paper is to validate the semi-AR framework, rather than to achieve state-of-the-art tokenizer performance.

\paragraph{Generator Training Details.}
We train decoder-only transformers on 17-frame videos with a resolution of 128$\times$128, using the UCF-101~\citep{Soomro2012UCF101AD} and K600~\citep{Carreira2018ASN} datasets. 
With spatial downsampling factors of $F_H=F_W=8$ and temporal downsampling of $F_T=4$, the resulting 3D token map for each video sample has dimensions $(T', H', W')=(5, 16, 16)$, yielding a total of 1280 tokens. 
We train our model for 100K steps with a total batch size of 256 and 64 respectively. 
Model sizes range from 700M to 3B parameters, with training spanning approximately two weeks on 32 NVIDIA A100 GPUs. The full model configuration and training hyper-parameters are provided in Appendix~\ref{app:model}. 
We train the models from scratch, rather than initializing from a pre-trained LLM checkpoint, as these text-based checkpoints provide minimal benefit for video generation~\citep{zhang2023pre}. 
We use LLaMA~\citep{Touvron2023LLaMAOA} vocabulary (32K tokens) as the text vocabulary and merge it with the video vocabulary (64K tokens) to form the final vocabulary. Since our primary focus is video generation, we compute the loss only on video tokens, which leads to improved performance. 

\input{tables/semiar_ar_scale}

\paragraph{Evaluation Protocol.}
We evaluate our models on the UCF-101 dataset for class-conditional generation task and the K600 dataset for frame prediction task. To assess video quality, we use the standard metric of Fréchet Video Distance (FVD)\cite{unterthiner2018towards}. Additional evaluation details can be found in Appendix\ref{app:eval}.



\begin{figure}[htbp]
    \centering
    \begin{minipage}[b]{0.49\linewidth}
        \centering
        \includegraphics[width=\linewidth]{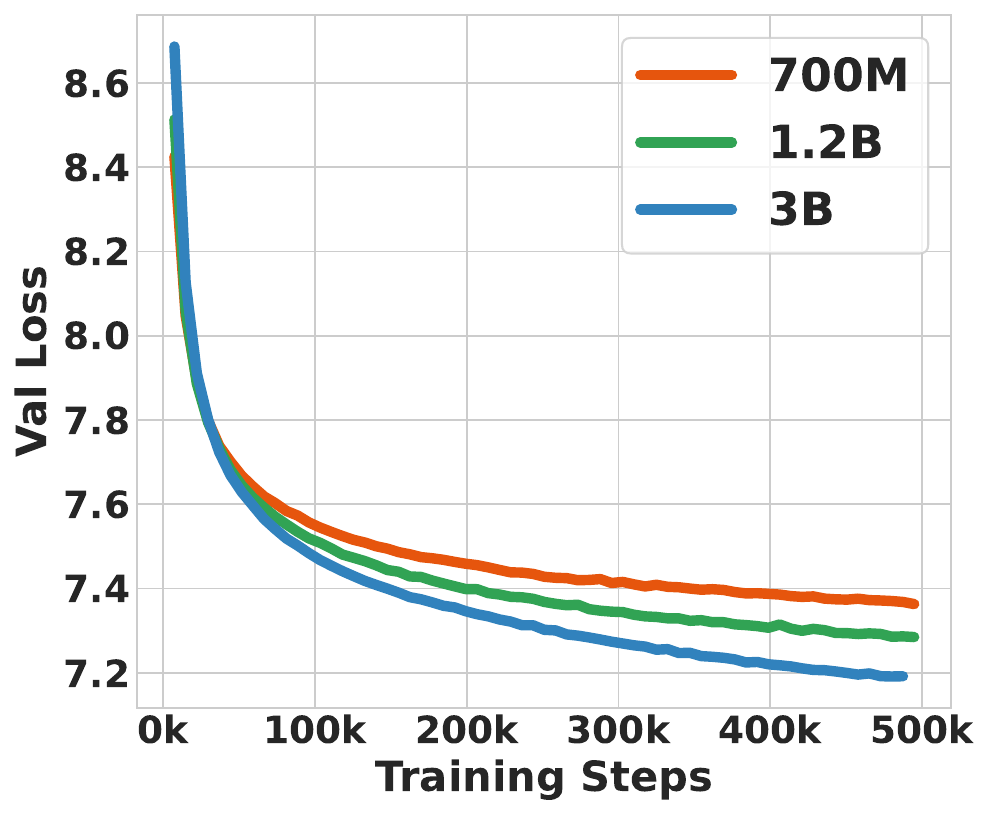}
        \caption{Validation loss of various sizes of semi-AR models from 700M to 3B. 
        }
\label{fig:model_para}
    \end{minipage}
    \hfill
    \begin{minipage}[b]{0.49\linewidth}
        \centering
        \includegraphics[width=\linewidth]{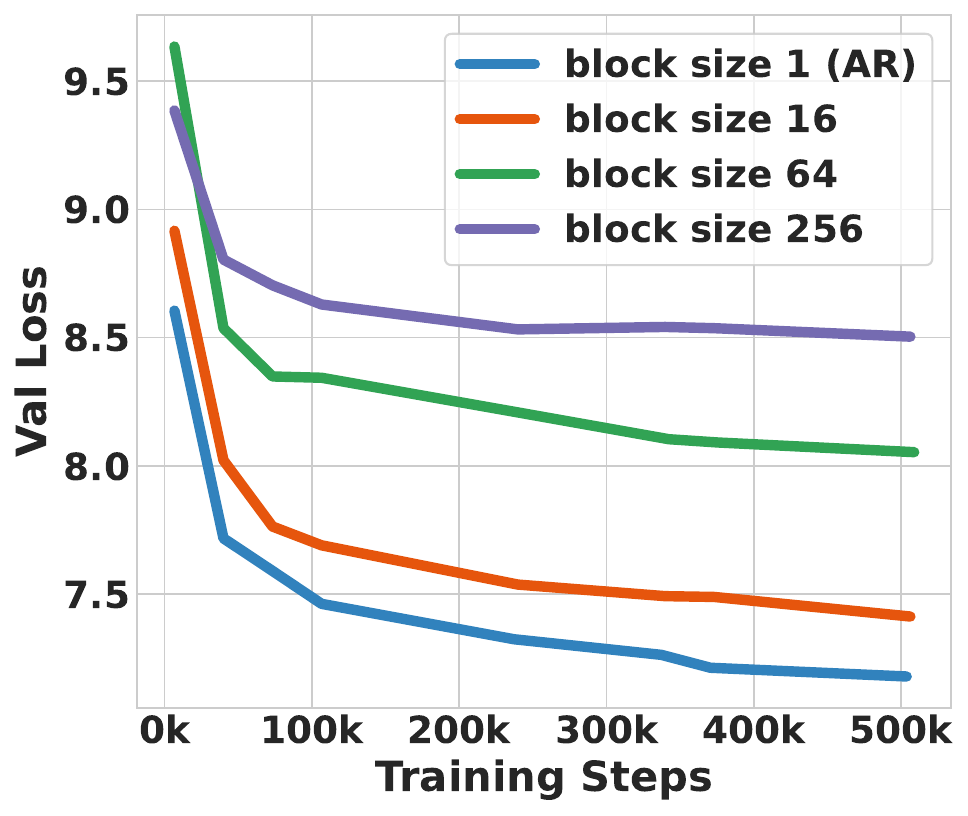}
        \caption{Validation loss of various block sizes from 1 to 256. }
        \label{fig:block_size}
    \end{minipage}
\end{figure}

\subsection{Comparison of Next-Token Prediction and Next-Block Prediction}

We first conduct a fair comparison between next-token prediction (NTP) and our next-block prediction (\modelname) under the same experimental setting. 
All experiments are performed on the K600 dataset, which has a much larger data volume compared to UCF-101 (413K vs. 9.5K) and features a strict training-test split, thereby ensuring more generalizable results. 
Table~\ref{table:semiar-ar-scale} highlights the superiority of our approach in three key aspects: generation quality, inference efficiency, and scalability.

\paragraph{Generation Quality.}
Across all model sizes, \modelname with a 1$\times$1$\times$16 block size consistently outperforms NTP models in terms of generation quality (measured by FVD). For instance, the 700M \modelname model achieves an FVD of 33.6, outperforming the NTP model by 3.8 points. Furthermore, a \modelname model with only 1.2B parameters achieves a comparable performance to a 3B NTP model (28.6 vs. 29.0 FVD). This suggests that the block size of 1$\times$1$\times$16 is a more effective generation unit for autoregressive modeling in the video domain. 

\input{tables/main}

\paragraph{Inference Efficiency.}
To generate a 12-frame video (128$\times$128 resolution, 768 tokens), a 700M NTP model requires 768 forward steps during inference, taking 15.04 seconds (FPS=0.80). 
In contrast, our \modelname model with a 1$\times$1$\times$16 block size predicts all tokens in a row simultaneously, requiring only 48 steps and 1.35 seconds to generate the video (FPS=8.89)—over 11 times faster than the NTP model. 
Since \modelname modifies only the target output and attention mask, it is compatible with the most efficient AR inference frameworks, such as memory-efficient attention~\citep{xFormers2022}, offering the potential for further speed improvements. 
We now briefly discuss the sources of these efficiency gains. 
In scenarios utilizing KV-Cache, the overall computation cost during each inference step for NTP involves multiplying vectors (current token) with matrices (model weights), which is primarily \textbf{IO-bound} due to the movement of matrices. Conversely, in the NBP model, the computation involves multiplying matrices (current block) with matrices (model weights), making it \textbf{compute-bound}, with reduced IO overhead due to larger block sizes. Given this distinction and assuming adequate GPU parallelism, the NBP framework can achieve significantly faster speeds compared to NTP. This efficiency gain is due to the reduced frequency of IO operations and the more effective utilization of computational resources in processing larger data blocks simultaneously.

\paragraph{Scalability.}
As model size increases from 700M to 1.2B and 3B parameters, the FVD of \modelname models improves from 33.6 to 28.6 and 26.5, respectively. This demonstrates that \modelname exhibits similar scalability to NTP models, with the potential for even greater performance as model size and computational resources increase. Fig.~\ref{fig:model_para} and Fig.~\ref{fig:vary_size_gen} present the validation loss curves and generation examples for different model sizes, respectively. 
As the models grow larger, the generated content exhibits greater stability and enhanced visual detail. 

\subsection{Benchmarking with Previous Systems}
\label{subsec:benchmark}
Table~\ref{tab:video_syn} presents our model's performance compared to strong baselines using various modeling approaches, including GAN, diffusion, mask token modeling (MTM), and vanilla AR methods. 
For UCF-101, the evaluation task is class-conditional video generation, where models generate videos based on a given class name. 
Our Semi-AR model, with 3B parameters, achieves an FVD of 55.3, surpassing HPDM-L~\citep{skorokhodov2024hierarchical} and MAGVITv2~\cite{yu2023language} by 11 and 2.7 FVD points, respectively.

For K600, the evaluation task is frame prediction, where all models predict future frames based on the same 5-frame condition from the validation set. Our 700M model achieves an FVD of 25.5, outperforming the strongest AR baseline, OmniTokenizer, by 7.4 FVD points.
While our model exhibits a performance gap compared to MAGVITv2, it achieves this result with significantly lower training computation (e.g., 77 epochs vs. MAGVITv2's 360 epochs). Scaling up the model size narrows this gap, with a 6-point improvement in FVD observed. Given the strong scalability of our semi-AR framework, we believe that with larger model sizes and increased training volumes, our approach could surpass MAGVITv2, akin to how large language models (LLMs)~\citep{Brown2020LanguageMA} have outperformed BERT~\citep{devlin2018bert} in NLP.

\subsection{Visualizations}
\paragraph{Video Reconstruction.}
Fig.~\ref{fig:vis_recons} compares the video reconstruction results of OmniTokenizer~\citep{Wang2024OmniTokenizerAJ} and our tokenizer. Our method significantly outperforms the baseline in both image clarity and motion stability.

\begin{figure*}[tbp]
\centering
\includegraphics[width=.9\textwidth]{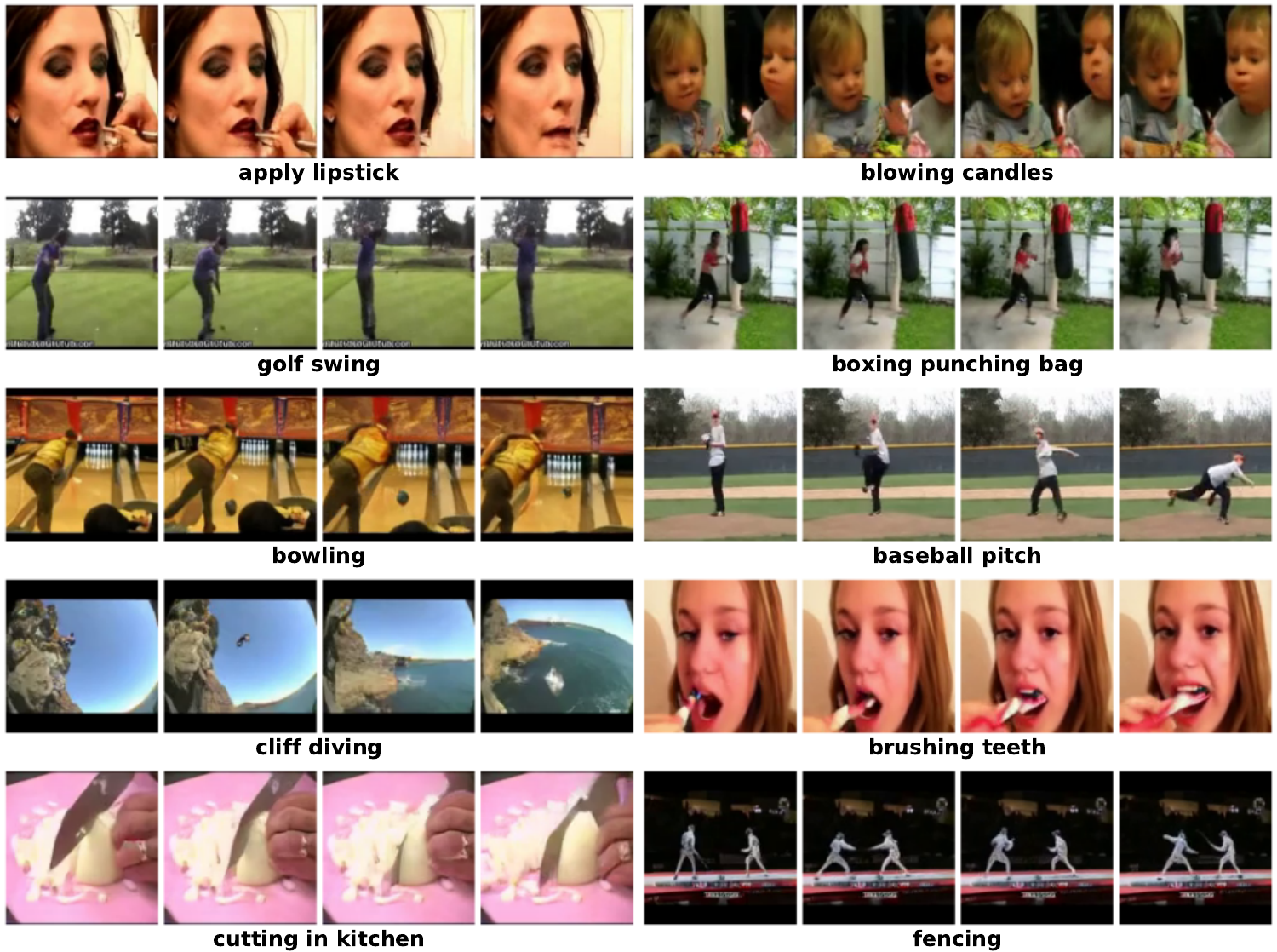}
\caption{Visualization of class-conditional generation (UCF-101) results of our method. The text below each video clip is the class name.}
\label{fig:ucf_gen}
\end{figure*}

\begin{figure*}[tbp]
\centering
\includegraphics[width=.9\textwidth]{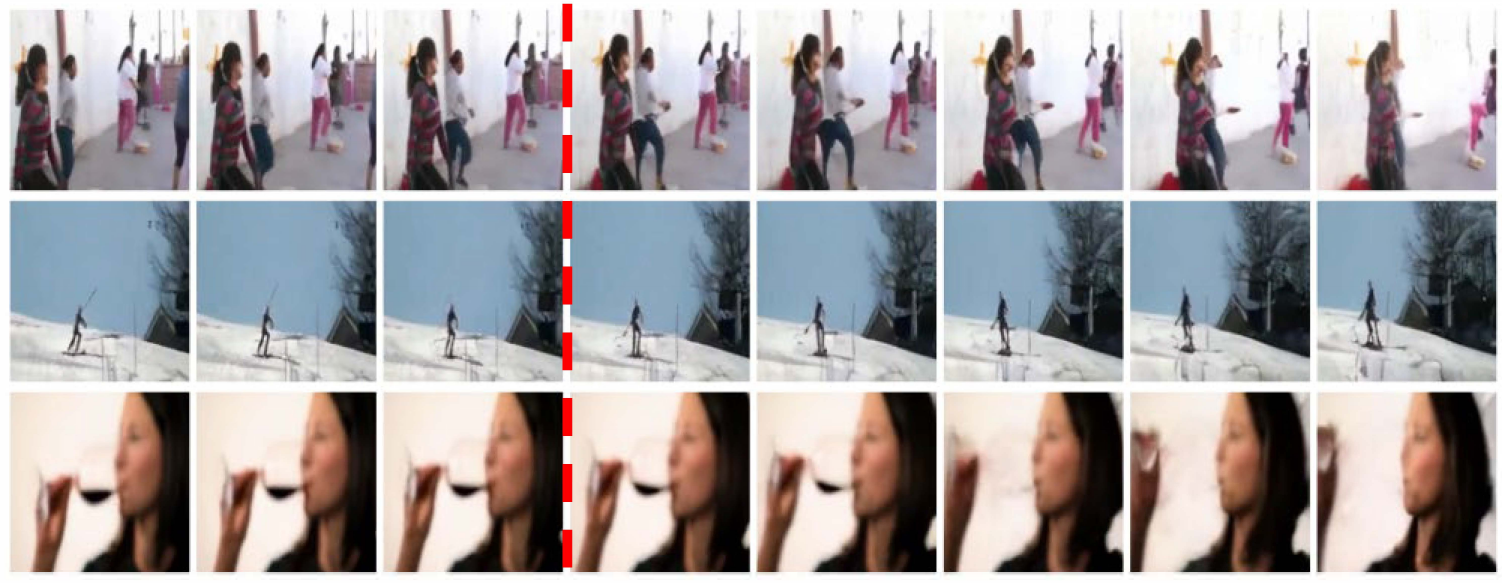}
\caption{Visualization of frame prediction (K600) results of our method.}
\label{fig:our_gen}
\end{figure*}

\begin{figure*}[tbp]
\centering
\includegraphics[width=.9\textwidth]{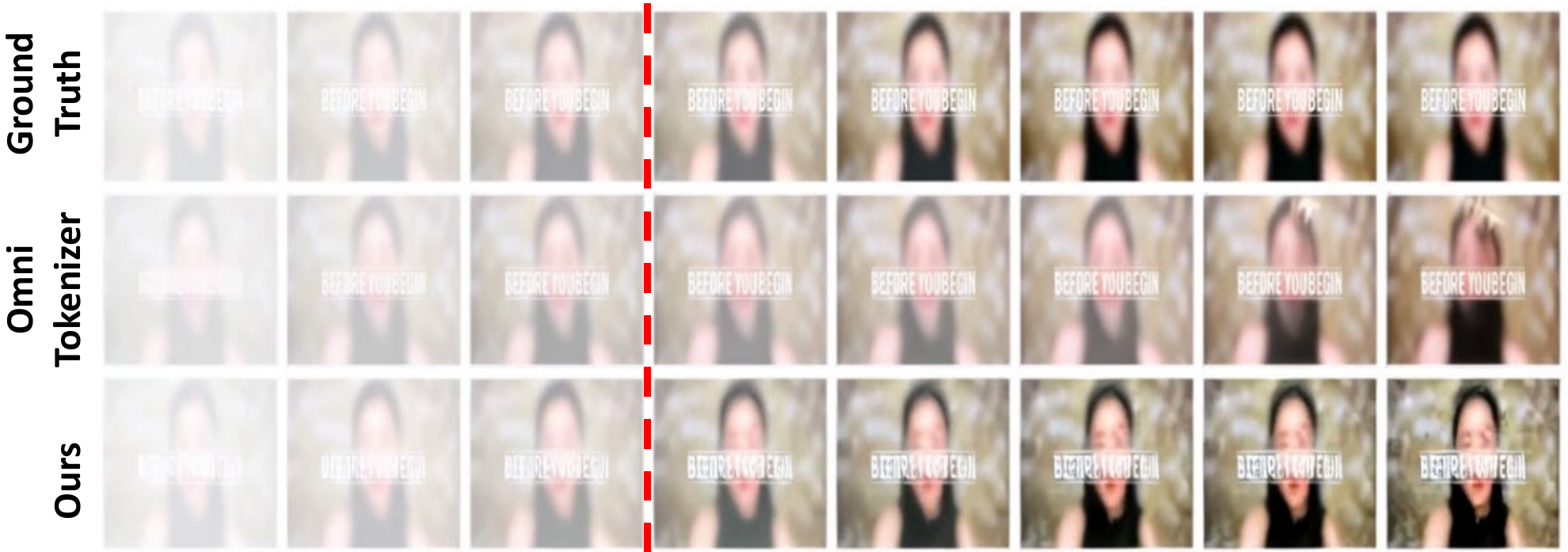}
\caption{Frame prediction results of OmniTokenizer and our method. The left part is the condition, and the right part is the predicted subsequent sequence.}
\label{fig:vis_gen}
\end{figure*}

\paragraph{Video Generation.}
The class-conditional generation results for UCF-101 are shown in Fig.\ref{fig:ucf_gen}, while the frame prediction results for K600 are shown in Figs.\ref{fig:our_gen}-\ref{fig:vis_gen}. The visualizations demonstrate that our model accurately predicts subsequent frames with high clarity and temporal coherence, even in scenarios involving large motion dynamics. 


\subsection{Ablation Study and Analysis}
\label{subsec:ablation}
In this subsection, we conduct an ablation study on block size and block shape, then analyze the attention patterns in our \modelname models.

\begin{figure}[htbp]
\includegraphics[width=\linewidth]{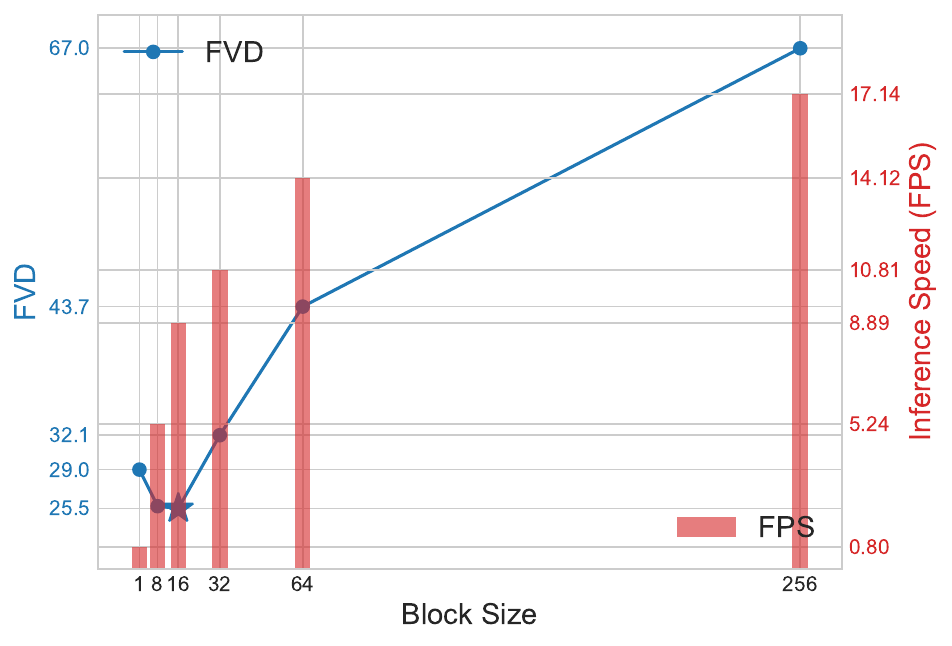}
\caption{Generation quality (FVD, lower is better) and inference speed (FPS, higher is better) of various block sizes from 1 to 256.}
\label{fig:block_size_fvd_fps}
\end{figure}

\paragraph{Ablation Study on Block Size.}
We experiment with different block sizes, ranging from $[1, 8, 16, 32, 64, 256]$\footnote{The full 3D size of the blocks are 1$\times$1$\times$1, 1$\times$1$\times$8, 1$\times$1$\times$16, 1$\times$2$\times$16, 1$\times$4$\times$16, 1$\times$16$\times 16$, respectively.}, to evaluate their impact on model performance. A block size of 1, 16, and 256 corresponds to token-by-token (NTP), row-by-row, and clip-by-clip generation, respectively. 
Fig.~\ref{fig:block_size} shows the validation loss curves for various block sizes. As block size decreases, learning becomes easier due to the increased prefix conditioning, which simplifies the prediction task and results in lower validation loss. 
However, due to the exposure bias associated with (semi-)AR modeling~\citep{ranzato2015sequence}, validation loss under the teacher-forcing setting does not completely correlate with final performance during inference~\citep{deng2024causal}. Notably, the smallest block size (i.e., a single token) does not yield optimal performance. As shown in Fig.~\ref{fig:block_size_fvd_fps}, a block size of 16 achieves the best generation quality, with an FVD improvement of 3.5 points, reaching 25.5. 
Block size is critical for balancing generation quality (FVD) and efficiency (FPS). While larger blocks (e.g., 1$\times$16$\times$16) lead to faster inference speeds (up to 17.14 FPS), performance degrades, indicating that generating an entire clip in one step is overly challenging. 
Additionally, inference decoding methods significantly influence results. As demonstrated in Fig.~\ref{fig:vary_block_gen}, traditional Top-P Top-K decoding can lead to screen fluctuations~\citep{lezama2022improved}, as it struggles to model spatial dependencies within large blocks, highlighting the need for improved decoding strategies in \modelname scenarios. 

\paragraph{Ablation Study on Block Shape.}
We explore the performance of various block shapes on K600, using the 700M model, the results are shown in Table~\ref{tab:block_shape}. 
Our findings indicate that the official block shape of T$\times$H$\times$W=1$\times$1$\times$16 (generating row by row) outperforms other tested shapes such as 1$\times$4$\times$4 and 2$\times$1$\times$8. We attribute this to two main factors: 
\textbf{(1) Token Relationships within a Single Block}: The shape of the 1$\times$1$\times$16 block allows tokens within the block to represent a complete, continuous row, maintaining integrity without cross-row interruptions. In contrast, block shapes like 1$\times$4$\times$4 and 2$\times$1$\times$8 involve generating complex relationships across multiple rows and columns—or even frames—on a smaller spatial scale, posing greater challenges~\citep{ren2023testa}. 
\textbf{(2) Relationships between Blocks}: The 1$\times$1$\times$16 block shape simplifies the modeling process to primarily vertical relationships between rows, which enhances continuity and consistency during generation, thereby reducing breaks and error accumulation.

\input{tables/block_shape}

\paragraph{Analysis of Attention Pattern.} 
We analyze the attention pattern in our \modelname framework using an example of next-clip prediction, where each block corresponds to a clip. 
Fig.~\ref{fig:txt_2clips_attn} shows the attention weights on UCF-101. Unlike the lower triangular distribution observed in AR models, our attention is characterized by a staircase pattern across blocks. In addition to high attention scores along the diagonal, the map reveals vertical stripe-like highlighted patterns, indicating that tokens at certain positions receive attention from all tokens. 
Fig.~\ref{fig:spatial-attn} illustrates the spatial attention distribution for a specific query (marked by \textcolor{red}{red $\times$}). This query can attend to all tokens within the clip, rather than being restricted to only the preceding tokens in a raster-scan order, enabling more effective spatial dependency modeling.

\section{Conclusion}
In this paper, we introduced a novel approach to video generation called Next Block Prediction using a semi-autoregressive modeling framework. This framework offers a more efficient and scalable solution for video generation, combining the advantages of parallelization with improved spatial-temporal dependency modeling. This method not only accelerates inference but also maintains or improves the quality of generated content, demonstrating strong potential for future applications in multimodal AI.




\section*{Impact Statement}
This work advances the field of video generation through the development of \modelname. While recognizing the potential of this technology, we carefully consider its societal implications, particularly regarding potential misuse and ethical challenges. The model's capabilities could be exploited to create harmful content, including deepfakes for misinformation campaigns or other malicious purposes. Furthermore, we acknowledge the critical importance of ensuring that generated content adheres to ethical standards by avoiding the perpetuation of harmful stereotypes and respecting cultural diversity. 
To mitigate these risks, we will explore a comprehensive framework of safeguards, including (1) robust digital watermarking to ensure traceability and accountability of generated content; (2) reinforcement learning with human feedback to align model outputs with ethical guidelines and reduce potential harm; and (3) clear usage policies and restrictions. These measures collectively aim to promote responsible development and deployment of video generation technology while maximizing its positive societal impact.

\nocite{langley00}

\bibliography{example_paper}
\bibliographystyle{icml2025}

\newpage
\appendix
\onecolumn
\section{Implementation Details}

\subsection{Task Definitions}
\label{app:task}
We introduce the tasks used in our training and evaluation. Each task is characterized by a few adjustable settings such as interior condition shape and optionally prefix condition. 
Given a video of shape $T\times H \times W$, we define the tasks as following:
\begin{itemize}
    \item Class-conditional Generation (CG)
    
    \begin{itemize}
        \item Prefix condition: class label.
    \end{itemize}
    
    

    \item Frame Prediction (FP)
    
    \begin{itemize}
        \item Interior condition: $t$ frames at the beginning; $t=5$ for K600 dataset.
    \end{itemize}
    
\end{itemize}
As we stated in $\S$~\ref{subsec:benchmark}, for UCF-101, all methods perform the CG task, while for K600, all methods perform the FP task.

\subsection{Model Configuration}
\label{app:model}

\paragraph{Video Tokenizer.}
Our video tokenizer shares the same model architecture with MAGVITv2~\cite{yu2023language}.

\paragraph{Decoder-only Generator.}
Table~\ref{tab:model_config} shows the configuration for the decoder-only generator. We use separate position encoding for text and video.  
We do not use advanced techniques in large language models, such as rotary position embedding (RoPE)~\citep{su2024roformer}, SwiGLU MLP, or RMS Norm~\citep{Touvron2023LLaMAOA}, which we believe could bring better performance.

\subsection{Training}
\paragraph{Video Tokenizer.}
Table~\ref{tab:tok_train_config} shows the training configurations of our video tokenizer. 

\paragraph{Decoder-only Generator.}
Table~\ref{tab:gen_train_config} shows the training configurations of our video generator.

For both tokenizer and generator training, the video samples are all 17 frames, frame stride 1, 128$\times$128 resolution.

\subsection{Evaluation}
\label{app:eval}

\paragraph{Evaluation metrics.}
The FVD~\cite{unterthiner2018towards} is used as the primary evaluation metric. 
We follow the official implementation\footnote{\url{https://github.com/google-research/google-research/tree/master/frechet_video_distance}} in extracting video features with an I3D model trained on Kinetics-400~\cite{carreira2017quo}.

\paragraph{Sampling protocols.}
We follow the sampling protocols from previous works~\cite{yu2023language, ge2022long,clark2019adversarial} when eveluating on the standard benchmarks, i.e. UCF-101, and Kinetics-600.
We sample 17-frame clips from each dataset without replacement to form the real distribution in FVD and extract condition inputs from them to feed to the model.
We continuously run through all the samples required (e.g., 40,000 for UCF-101) with a single data loader and compute the mean and standard deviation for 4 folds. 
We use top-$p$ and top-$k$ sampling with $k=16,000$ and $p=0.9$. 

Below are detailed setups for each dataset:

\begin{itemize}
    \item UCF-101: 
    \begin{itemize}
        \item Dataset: 9.5K videos for training, 101 classes.
        \item Number of samples: 10,000$\times$4.
        \item Resolution: 128$\times$128.
        \item Real distribution: random clips from the training videos.
        \item Video FPS: 8.
    \end{itemize}
    \item Kinetics-600:
    \begin{itemize}
        \item Dataset: 384K videos for training and 29K videos for evaluation.
        \item Number of samples: 50,000$\times$4.
        \item Generation resolution: 128$\times$128.
        \item Evaluation resolution: 64$\times$64, via central crop and bilinear resize.
        \item Video FPS: 25.
    \end{itemize}
\end{itemize}

\input{tables/model_config}
\input{tables/tok_train_config}
\input{tables/gen_train_config}

\section{Performance of Video Tokenizer}
\label{sec:vid-tok}
\input{tables/tokenizer}

\begin{figure*}[tbp]
\centering
\includegraphics[width=\textwidth]{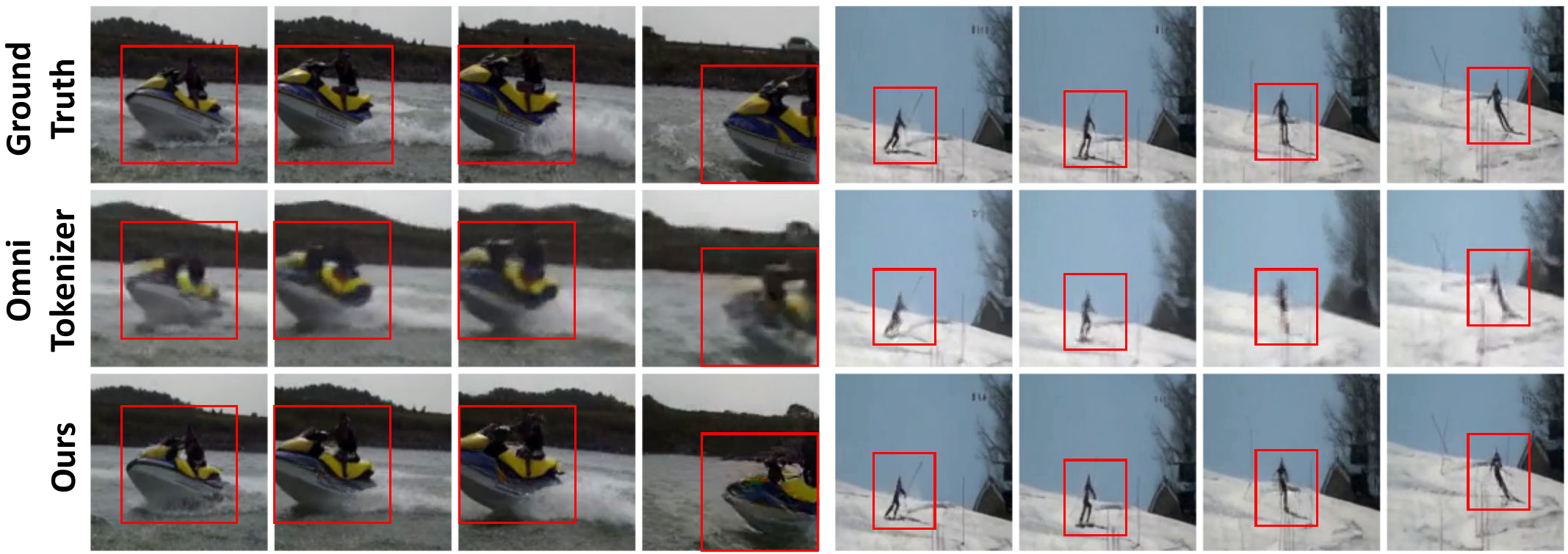}
\caption{Video reconstruction results (17 frames 128$\times$128 resolution at 25 fps and shown at 6.25 fps) of OmniTokenizer and our method. }
\label{fig:vis_recons}
\end{figure*}

We present the reconstruction performance of our tokenizer in Table~\ref{tab:video_reconstruction}. Our tokenizer achieves 15.50 rFVD on UCF-101 and 6.73 rFVD on K600, surpassing OmniTokenizer~\cite{Wang2024OmniTokenizerAJ}, MAGVITv1~\cite{yu2023magvit}, and other models. Fig.~\ref{fig:vis_recons} compares the video reconstruction results of OmniTokenizer~\citep{Wang2024OmniTokenizerAJ} and our tokenizer. Our method significantly outperforms the baseline in both image clarity and motion stability. 

\section{Visualization}
\begin{figure*}[tbp]
\centering
\includegraphics[width=\textwidth]{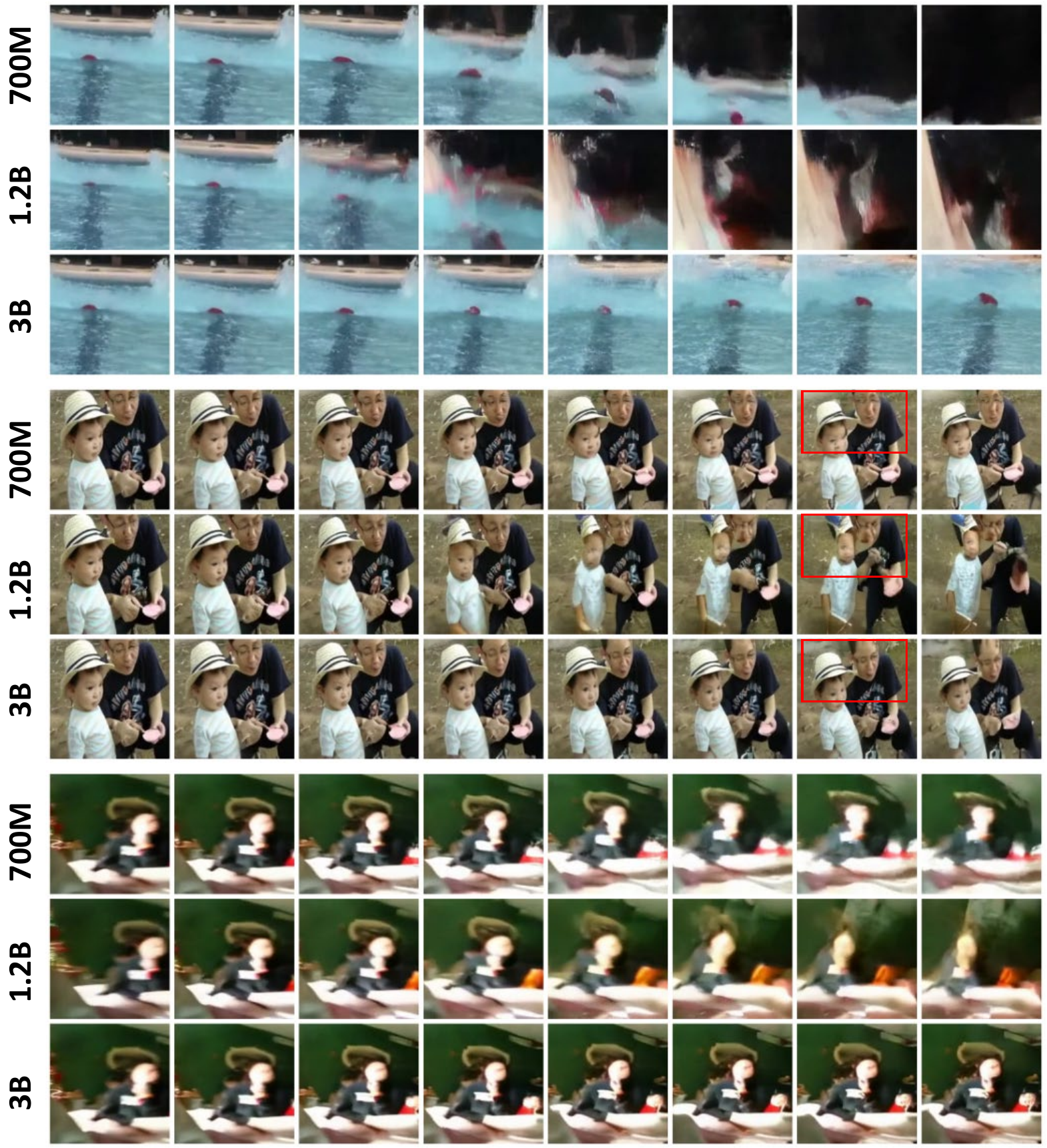}
\caption{Visualization of video generation results of various model sizes (700M, 1.2B, and 3B).}
\label{fig:vary_size_gen}
\end{figure*}

\begin{figure*}[tbp]
\centering
\includegraphics[width=\textwidth]{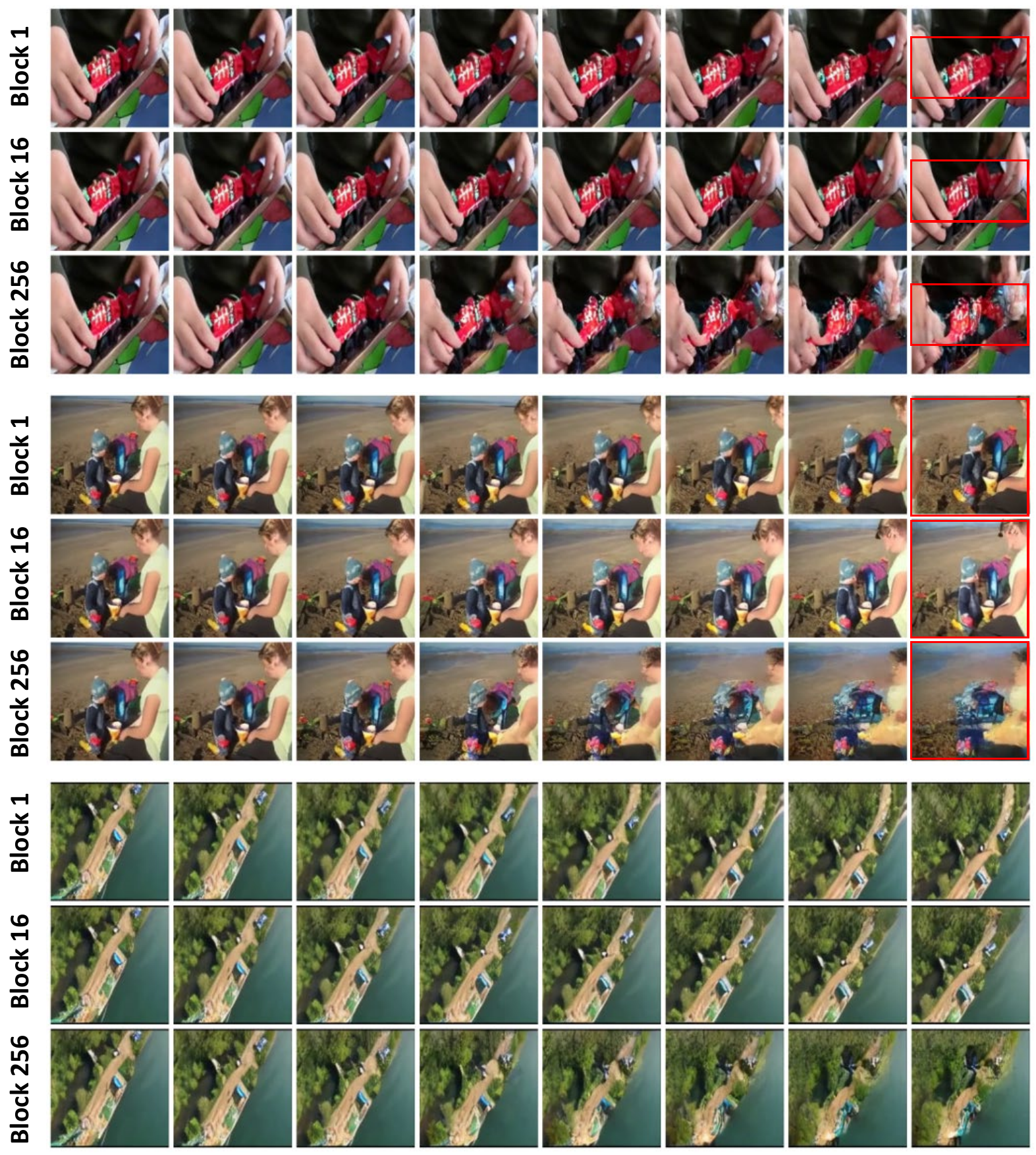}
\caption{Visualization of video generation results of various block sizes (1$\times$1$\times$1, 1$\times$1$\times$16 and 1$\times$16$\times$16).}
\label{fig:vary_block_gen}
\end{figure*}

We provide additional visualization of video generation results. 
Fig.~\ref{fig:vary_size_gen} shows results of various model sizes (700M, 1.2B and 3B).
Fig.~\ref{fig:vary_block_gen} shows results of various block sizes (1$\times$1$\times$1, 1$\times$1$\times$16 and 1$\times$16$\times$16).

\begin{figure*}[tbp]
\centering
\includegraphics[width=.9\textwidth]{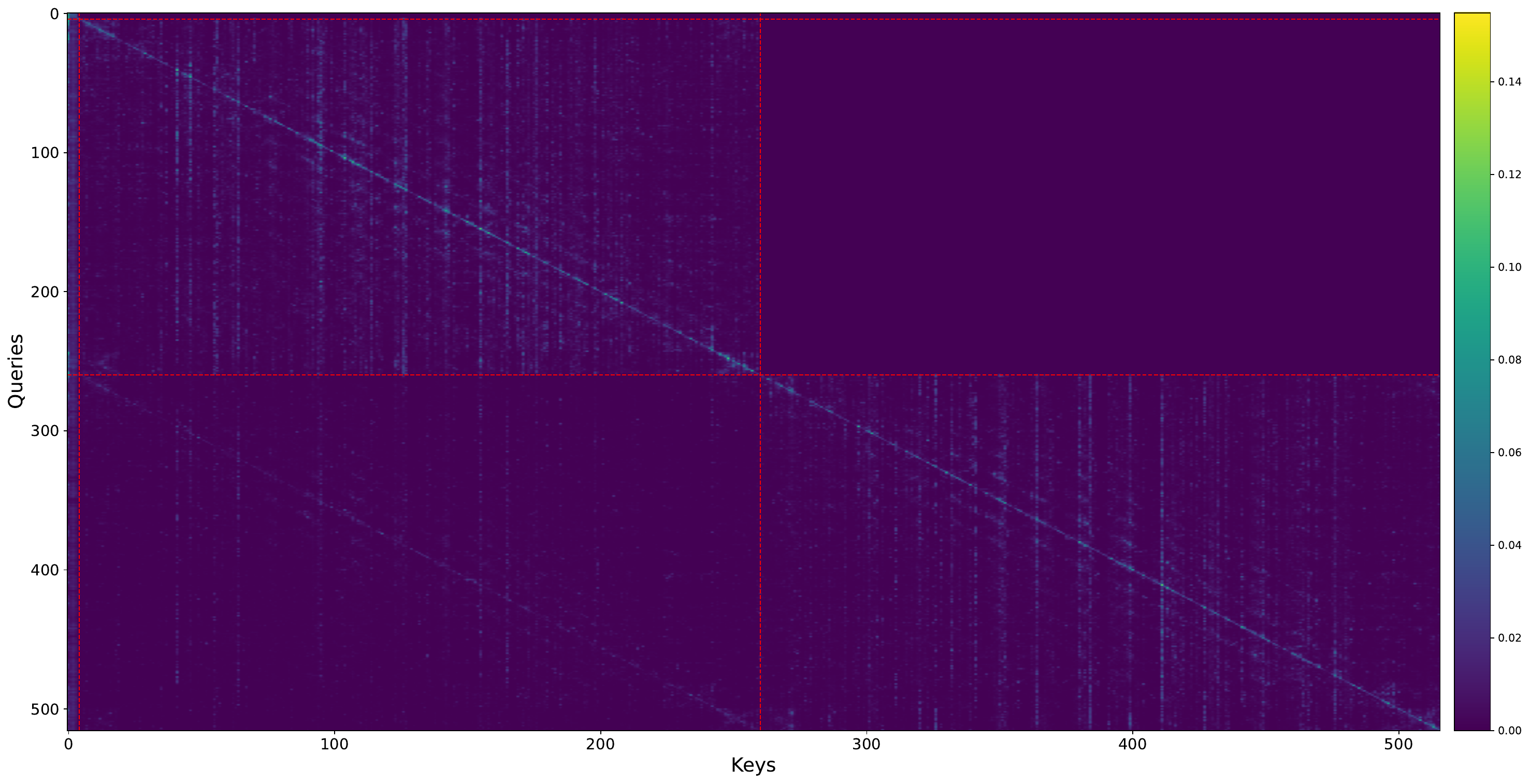}
\caption{Attention weights of next-clip prediction on UCF-101. The horizontal and vertical axis represent the keys and queries, respectively. Two red lines on each axis divide the axis into three segments, corresponding to the text (classname), the first clip, and the second clip. The brightness of each pixel reflects the attention score. We downweight the attention to text tokens by $5\times$ to provide a more clear visualization.}
\label{fig:txt_2clips_attn}
\end{figure*}

\begin{figure*}[tbp]
\centering
\includegraphics[width=.9\textwidth]{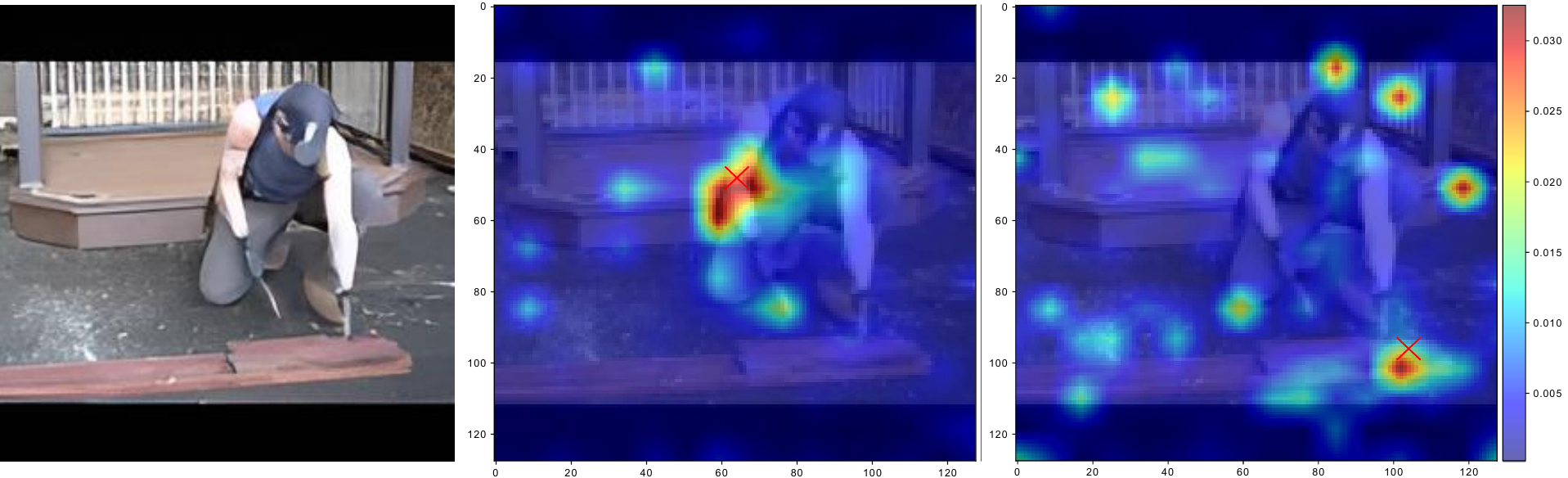}
\caption{Spatial attention distribution for a specific query (represented by \textcolor{red}{red $\times$}) on UCF-101.}
\label{fig:spatial-attn}
\end{figure*}

\end{document}

%% file: tables/semiar_ar_scale.tex
\begin{table*}[]
\centering
\caption{Comparison of next-token prediction (NTP) and next-block prediction (\modelname) models in terms of performance and speed, evaluated on the K600 dataset (5-frame condition, 12 frames (768 tokens) to predict). Inference time was measured on a single A100 Nvidia GPU. All models are implemented by us under the same setting and trained for 20 epochs. FPS denotes ``frame per second''. The measurement of inference speed includes tokenization and de-tokenization processes. KV-cache is used for both models.}
\label{table:semiar-ar-scale}
\begin{adjustbox}{max width=\linewidth}

\begin{tabular}{@{}c|l|c|c|cc@{}}
\toprule
Model Size & Modeling Method & \# Block size & FVD $\downarrow$ & \# Forward steps & Inference speed (FPS) $\uparrow$ \\ \midrule
\multirow{2}{*}{700M}          & NTP     & 1 (1$\times$1$\times$1)   & 37.4 & 768           & 0.80                  \\
                               & \modelname (Ours)  & 16 (1$\times$1$\times$16) & \textbf{33.6} & \textbf{48}            & \textbf{8.89}                  \\ \midrule
\multirow{2}{*}{1.2B}          & NTP  & 1 (1$\times$1$\times$1)    & 31.4 & 768           & 0.75                  \\
                               & \modelname (Ours)  & 16 (1$\times$1$\times$16) & \textbf{28.6} & \textbf{48}            & \textbf{6.70}                  \\ \midrule
\multirow{2}{*}{3B}            & NTP  & 1 (1$\times$1$\times$1)    & 29.0 & 768           & 0.60                  \\
                               & \modelname (Ours)  & 16 (1$\times$1$\times$16)  & \textbf{26.5} & \textbf{48}            & \textbf{4.29}                  \\ \bottomrule
\end{tabular}

\end{adjustbox}
\end{table*}

%% file: tables/main.tex
\begin{table*}[t]
\caption{Comparions of class-conditional generation results on UCF-101 and frame prediction results on K600. MTM indicates mask token modeling. Our model on K600 is trained for 77 epochs, we gray out models that use significantly more training computation (e.g., those trained for over 300 epochs) for a fair comparison.}
\label{tab:video_syn}
\centering
\renewcommand{\arraystretch}{1.1}
\setlength{\tabcolsep}{2.0pt}
\vspace{0.04in}
\begin{adjustbox}{max width=.9\linewidth}

\begin{tabular}{lcc|cc|rrr|rrr}
\toprule
\multirow{2}*{Type}& \multirow{2}*{Method} && \multirow{2}*{$\#$Param} && \multicolumn{3}{c}{UCF-101} & \multicolumn{3}{|c}{K600} \\
~ & ~ && ~ && FVD$\downarrow$ & \# Token & \# Steps & FVD$\downarrow$ & \# Token & \# Steps \\ 
\midrule
GAN & 	DVD-GAN~\citep{clark2019adversarial} && N/A && - & - & -  & 31.1 & - & - \\
\midrule
Diffusion & VideoFusion~\citep{luo2023videofusion} && N/A && 173 & - & - & - & - & - \\
Diffusion & Make-A-Video~\citep{singer2022make} && N/A && 81.3 & - & - & - & - & - \\
Diffusion & HPDM-L~\citep{skorokhodov2024hierarchical} && 725M && 66.3 & - & - & - & - & - \\
\midrule
MTM & Phenaki~\cite{villegas2022phenaki} && 227M && - & - & - & \demph{36.4} & \demph{-} & \demph{48} \\
MTM & MAGVIT~\cite{yu2023magvit} && 306M && 76 & 1280 & 12 & \demph{9.9} & \demph{768} & \demph{12} \\
MTM & MAGVITv2~\cite{yu2023language} && 840M && 58 & 1280 & 24 & \demph{4.3} & \demph{768} & \demph{24} \\
\midrule
AR & LVT~\cite{rakhimov2020latent} && 50M && - & - & - & 224.7 & 1024 & 1024 \\
AR & ViTrans~\cite{weissenborn2019scaling} && 373M && - & - & - & 170.0 & 4096 & 4096 \\
AR & CogVideo~\cite{hong2022cogvideo} && 9.4B && 626 & 2000 & 2000 & 109.2 & 2000 & 2000 \\
AR & ViVQVAE~\cite{walker2021predicting} && N/A && - & - & - & 64.3 & 4096 & 4096 \\
AR & TATS~\cite{ge2022long} && 321M && 332 & 1024 & 1024 & - & - & - \\
AR & OmniTokenizer~\cite{Wang2024OmniTokenizerAJ} && 227M && 314 & 5120 & 5120 & \demph{34.2} & \demph{3072} & \demph{3072} \\
AR & OmniTokenizer~\cite{Wang2024OmniTokenizerAJ} && 650M && 191 & 5120 & 5120 & \demph{32.9} & \demph{3072} & \demph{3072} \\
AR & MAGVITv2-AR~\cite{yu2023language} && 840M && 109 & 1280 & 1280 & - & - & - \\
AR & PAR-16$\times$~\citep{wang2024parallelized} && 792M && 103.4 & 1280 & 95 & - & - & - \\
\midrule
Semi-AR & \modelname-XL (Ours) && 700M && 103.3 & 1280 & 95 & 25.5 & 768 & 48 \\
Semi-AR & \modelname-XXL (Ours) && 1.2B && 85.8 & 1280 & 95 & 23.0 & 768 & 48 \\
Semi-AR & \modelname-3B (Ours) && 3B && \textbf{55.3} & 1280 & 95 & \textbf{19.5} &  768 & 48 \\
\bottomrule
\end{tabular}

\end{adjustbox}
\end{table*}

%% file: tables/block_shape.tex
\begin{table}[htbp]
\centering
\caption{Generation quality (FVD) of various block shape.}
\label{tab:block_shape}
\begin{tabular}{ccc}
\toprule
Block Size & Block Shape (T$\times$H$\times$W) & FVD$\downarrow$ \\
\midrule
16 & 1$\times$4$\times$4 & 33.4 \\
16 & 2$\times$1$\times$8 & 29.2 \\ 
16 & 1$\times$1$\times$16 & \textbf{25.5} \\\midrule
8  & 2$\times$2$\times$2  & 32.7 \\
8  & 1$\times$1$\times$8  & \textbf{25.7} \\
\bottomrule
\end{tabular}
\end{table}

%% file: tables/model_config.tex
\begin{table*}
\caption{Model sizes and architecture configurations of our generation model. The configurations are following LLaMA~\citep{Touvron2023LLaMAOA}.
}
\centering
\begin{tabular}{@{}lcccc@{}}
\toprule
Model & Parameters & Layers & Hidden Size & Heads \\
\midrule
\modelname-XL & 700M & 24 & 1536 &  16 \\
\modelname-XXL & 1.2B & 24 & 2048 &  32 \\
\modelname-3B & 3B & 32 & 3072 &  32 \\
\bottomrule
\end{tabular}
\label{tab:model_config}
\end{table*}

%% file: tables/tok_train_config.tex
\begin{table}[]
\caption{Training configurations of video tokenizer.}
\label{tab:tok_train_config}
\centering
\begin{tabular}{l|cc}
\toprule
Hyper-parameters                  & UCF101                          & K600                            \\ \midrule
Video  FPS              & 8 & 8 \\
Latent shape                      & 5$\times$16$\times$16           & 5$\times$16$\times$16           \\
Vocabulary size                   & 64K                           & 64K                           \\
Embedding dimension               & 6                               & 6                               \\
Initialization                    & Random                          & Random                          \\
Peak learning rate                & 5e-5                            & 1e-4                            \\
Learning rate schedule            & linear & linear              \\
Warmup ratio                      & 0.01                            & 0.01                            \\
Perceptual loss weight            & 0.1                             & 0.1                             \\
Generator adversarial loss weight & 0.1                             & 0.1                             \\
Optimizer                         & Adam                            & Adam                            \\
Batch size                        & 256                             & 256                             \\
Epoch                             & 2000                            & 100                             \\ \bottomrule
\end{tabular}
\end{table}

%% file: tables/gen_train_config.tex
\begin{table}[]
\caption{Training configurations of video generator (base model).}
\label{tab:gen_train_config}
\centering
\begin{tabular}{l|cc}
\toprule
Hyper-parameters                  & UCF101                          & K600                            \\ \midrule
Video  FPS              &  8  &  16 \\
Latent shape                      & 5$\times$16$\times$16           & 5$\times$16$\times$16           \\
Vocabulary size                   & 96K (including 32K text tokens)                          & 64K                           \\
Initialization                    & Random                          & Random                          \\
Peak learning rate                & 6e-4                            & 1e-3                            \\
Learning rate schedule            & linear & linear              \\
Warmup steps                      & 5,000                             & 10,000                             \\
Weight decay & 0.01 & 0.01 \\
Optimizer                         & Adam (0.9, 0.98)                           & Adam (0.9, 0.98)                           \\
Dropout & 0.1 & 0.1 \\
Batch size                        & 256                             & 64                             \\
Epoch                             & 2560                            & 77                             \\ \bottomrule
\end{tabular}
\end{table}

%% file: tables/tokenizer.tex
\begin{table*}[]
  \centering
  \caption{Video reconstruction results on UCF-101 and K600.}
  \label{tab:video_reconstruction}
  \setlength{\tabcolsep}{2.0pt}
    \vspace{0.04in}
  \resizebox{\linewidth}{!}{
  \begin{tabular}{lccccrrrrrrrr}
    \toprule
    \multicolumn{5}{c}{} & \multicolumn{4}{c}{UCF-101} & \multicolumn{4}{c}{K600} \\
    \cmidrule(r){6-9} \cmidrule(r){10-13}
    Method & Backbone & Quantizer & Param. & \# bits& rFVD$\downarrow$  & PSNR$\uparrow$ & SSIM$\uparrow$ & LPIPS$\downarrow$ & rFVD$\downarrow$ & PSNR$\uparrow$ & SSIM$\uparrow$ & LPIPS$\downarrow$ \\
    \midrule
    MaskGIT~\cite{chang2022maskgit} & 2D CNN & VQ & 53M & 10 & 216 & 21.5 & .685 & .1140 & - & - & - & - \\
    TATS~\cite{ge2022long} & 3D CNN & VQ & 32M & 14 & 162 & - & - & - & - & - & - & - \\
    OmniTokenizer~\cite{Wang2024OmniTokenizerAJ} & ViT & VQ & 78M & 13 & 42 & 30.3 & .910 & .0733 & 27 & 28.5 & .883 & .0945 \\
    MAGVIT-v1~\cite{yu2023magvit} & 3D CNN & VQ & 158M & 10 & 25 & 22.0 & .701 & .0990 & - & - & - & - \\
    MAGVIT-v2~\cite{yu2023language} & C.-3D CNN & LFQ & 158M & 18 & 16.12 & - & - & .0694 & - & - & - & - \\
    MAGVIT-v2~\cite{yu2023language} & C.-3D CNN & LFQ & 370M & 18  & 8.62 & - & - & .0537 & - & - & - & - \\
    \midrule
    \modelname-Tokenizer (Ours) & C.-3D CNN & FSQ & 370M & 16 & 15.50 & 29.3 & .893 & .0648 & 6.73 & 31.3 & .944 & .0828 \\
    \bottomrule
  \end{tabular}
  }
  \vspace{-10pt}
\end{table*}